\documentclass{llncs} 

\usepackage{amsmath}
\usepackage{amssymb,mathtools} 
\usepackage{multirow}
\usepackage{tabularx,graphicx}
\usepackage{booktabs}
\usepackage[colorlinks]{hyperref}
\usepackage{color,soul}
\usepackage{siunitx}
\usepackage{subfigure}
\usepackage{lipsum}
\usepackage{caption}

\title{Learning to Map Vehicles into Bird's Eye View}

\author{Andrea Palazzi, Guido Borghi, Davide Abati,\\ Simone Calderara and Rita Cucchiara}

\institute{University of Modena and Reggio Emilia, Italy\\
         {\tt\small \{name.surname\}@unimore.it}
}

\begin{document}

\maketitle
\begin{abstract}
Awareness of the road scene is an essential component for both autonomous vehicles and Advances Driver Assistance Systems and is gaining importance both for the academia and car companies. 
This paper presents a way to learn a semantic-aware transformation which maps detections from a dashboard camera view onto a broader bird's eye occupancy map of the scene. To this end, a huge synthetic dataset featuring 1M couples of frames, taken from both car dashboard and bird's eye view, has been collected and automatically annotated. A deep-network is then trained to warp detections from the first to the second view. We demonstrate the effectiveness of our model against several baselines and observe that is able to generalize on real-world data despite having been trained solely on synthetic ones.
\end{abstract}
%
%
%
%
\section{Introduction} \label{sec:intro}
Vision-based algorithms and models have massively been adopted in current generation ADAS solutions. Moreover, recent research achievements on scene semantic segmentation \cite{semseg_ghiasi2016laplacian,semseg_lin2016efficient}, road obstacle detection \cite{obstacles_bernini2014real,obstacles_levi2015stixelnet} and driver's gaze, pose and attention prediction \cite{dong2011driver,ventu2016} are likely to play a major role in the rise of autonomous driving.\\
As suggested in \cite{synthetic_chen2015deepdriving}, three major paradigms can be individuated for vision-based autonomous driving systems: \textit{mediated perception} approaches, based on the total understanding of the scene around the car, \textit{behavior reflex} methods, in which driving action is regressed directly from the sensory input, and \textit{direct perception} techniques, that fuse elements of previous approaches and learn a mapping between the input image and a set of interpretable indicators which summarize the driving situation.\\
Following this last line of work, in this paper we develop a model for mapping vehicles across different views. In particular, our aim is to warp vehicles detected from a dashboard camera view into a bird's eye occupancy map of the surroundings, which is an easily interpretable proxy of the road state. Being almost impossible to collect a dataset with this kind of information in real-world, we exclusively rely on synthetic data for learning this projection.\\
We aim to create a system close to surround vision monitoring ones, also called around view cameras that can be useful tools for assisting drivers during maneuvers by, for example, performing trajectory analysis of vehicles out from own visual field.

\begin{figure*}[t!]
\begin{center}
\includegraphics[width=0.85\textwidth]{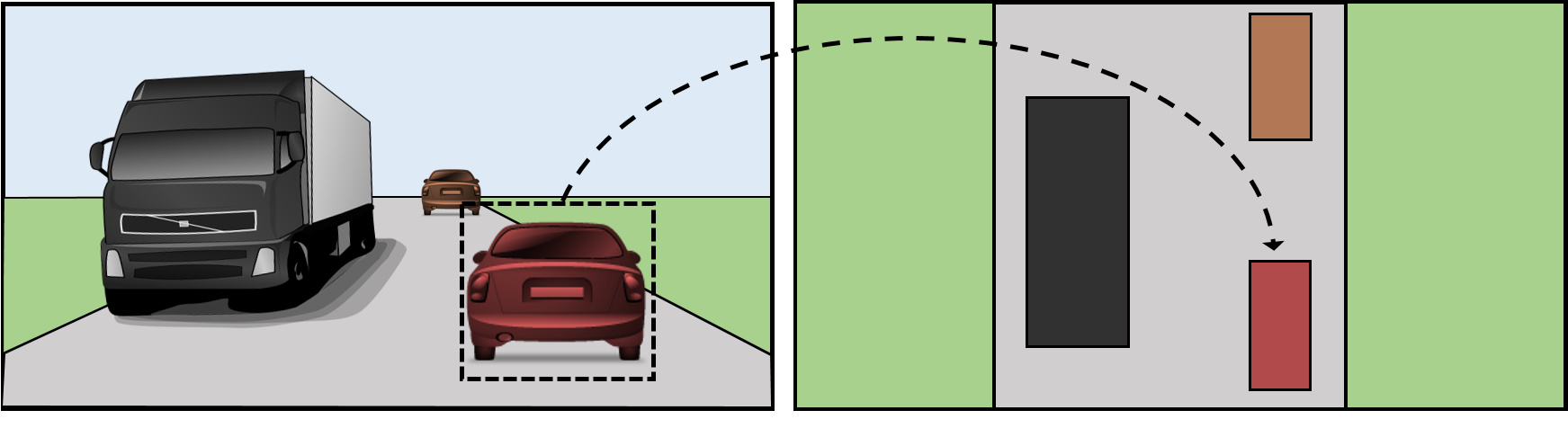}
\end{center}
   \caption{Simple outline of our task. Vehicle detections in the frontal view (left) are mapped onto a bird-eye view (right), accounting for the positions and size.}
\label{fig:task_overview}
\end{figure*}

\noindent In this framework, our contribution is twofold:
\begin{itemize}
\item We make available a huge synthetic dataset ($>$ 1 million of examples) which consists of couple of frames corresponding to the same driving scene captured by two different views. Besides the vehicle location, auxiliary information such as the distance and yaw of each vehicle at each frame are also present.
\item We propose a deep learning architecture for generating bird's eye occupancy maps of the surround in the context of autonomous and assisted driving. Our approach does not require a stereo camera, nor more sophisticated sensors like radar and lidar. Conversely, we learn how to project detections from the dashboard camera view onto a broader bird's eye view of the scene (see Fig.\ref{fig:task_overview}). To this aim we combine learned geometric transformation and visual cues that preserve objects size and orientation in the warping procedure.
\end{itemize}
Dataset, code and pre-trained model are publicly available and can be found at \url{http://imagelab.ing.unimore.it/scene-awareness}.
%
%
%
%
\section{Related work} \label{sec:related}
\subsubsection{Surround view}
Few works in literature tackle the problem of the vehicle's surround view. Most of these approaches are vision and geometry based and are specifically tailored for helping drivers during parking manoeuvres. 
In particular, in \cite{lin2012} a perspective projection image is transformed into its corresponding bird's eye view, through a fitting parameters searching algorithm. 
In \cite{liu2008} exploited the calibration of six fish eye cameras to integrate six images into a single one, by a dynamic programming approach. 
In \cite{nielsen2005} were described algorithms for creating, storing and viewing surround images, thanks to synchronized and aligned different cameras. Sung \textit{et al.} \cite{sung2012} proposed a camera model based algorithm to reconstruct and view multi-camera images. 
In \cite{tseng2013}, an homography matrix is used to perform a coordinate transformation: visible markers are required in input images during the camera calibration process. \\
Recently, Zhang \textit{et al.} \cite{zhang2014} proposed a surround view camera solution designed for embedded systems, based on a geometric alignment, to correct lens distortions, a photometric alignment, to correct brightness and color mismatch and a composite view synthesis. 
\subsubsection{Videgames for collecting data} The use of synthetic data has recently gained considerable importance in the computer vision community for several reasons. First, modern open-world games exhibit constantly increasing realism - which does not only mean that they feature photorealistic lights/textures etc, but also show plausible game dynamics and lifelike autonomous entity AI~\cite{synthetic_richter2016playing,synthetic_ros2016synthia}~. Furthermore, most research fields in computer vision are now tackled by means of deep networks, which are notoriously data hungry in order to be properly trained.
Particularly in the context of assisted and autonomous driving, the opportunity to exploit virtual yet realistic worlds for developing new techniques has been embraced widely: indeed, this makes possible to postpone the (very expensive) validation in real world to the moment in which a new algorithm already performs reasonably well in the simulated environment \cite{synthetic_wymann2000torcs,synthetic_gaidon2016virtual}. Building upon this tendency, \cite{synthetic_chen2015deepdriving} relies on TORCS simulator to learn an interpretable representation of the scene useful for the task of autonomous driving. However, while TORCS~\cite{synthetic_wymann2000torcs} is a powerful simulation tool, it's still severely limited by the fact that both its graphics and its game variety and dynamics are far from being realistic.\\
\\
Many elements mark as original our approach. In principle, we want our surround view to include not only nearby elements, like commercial geometry-based systems, but also most of the elements detected into the acquired dashboard camera frame. Additionally, no specific initialization or alignment procedures are necessary: in particular, no camera calibration and no visible alignment points are required. Eventually, we aim to preserve the correct dimensions of detected objects, which shape is mapped onto the surround view consistently with their semantic class.  

\begin{figure*}[t]
    \centering
    \begin{tabular}{c}
        \hspace{-0.0cm}\includegraphics[width=\textwidth]{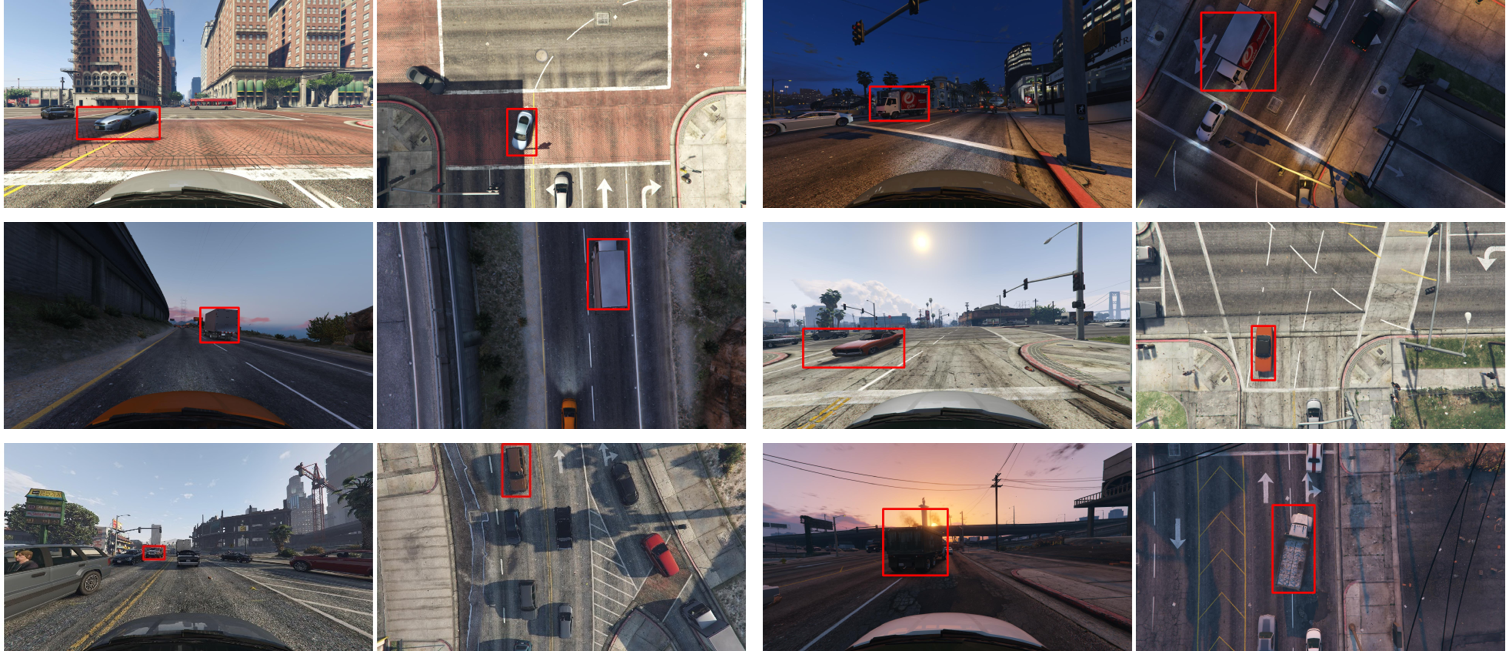}\\
        (a)\\
        \hspace{-0.0cm}\includegraphics[width=\textwidth]{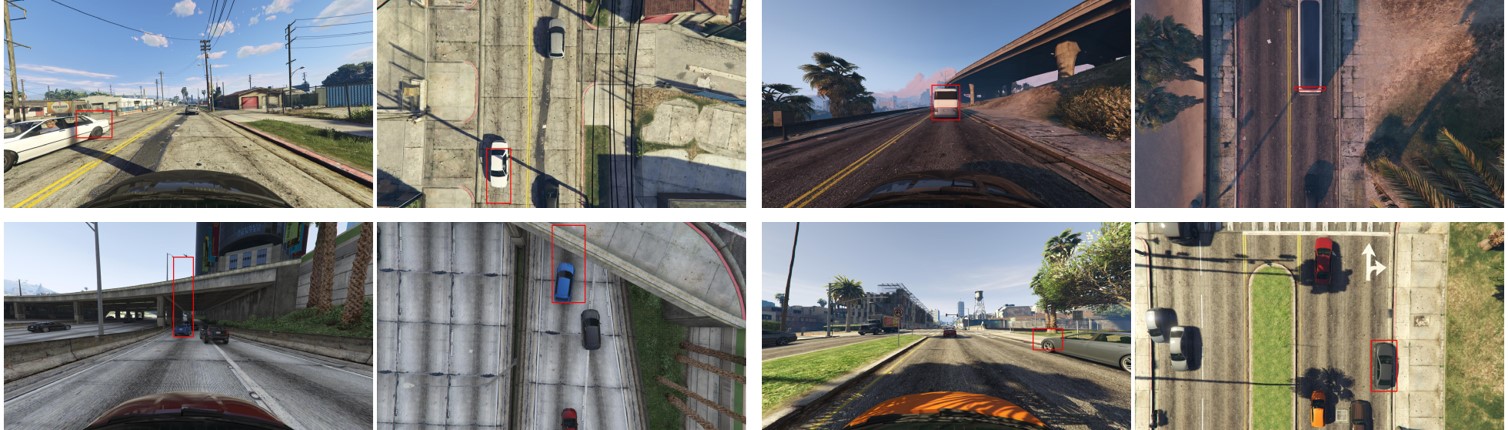}\\
        (b)\\
    \end{tabular}
    \caption{(a) Randomly sampled couples from our GTAV dataset, which highlight the huge variety in terms of landscape, traffic condition, vehicle models etc. Each detection is treated as a separate training example (see Sec.~\ref{sec:dataset} for details). (b)~Random examples rejected during the post-processing phase.}
    \label{fig:dataset_samples}
\end{figure*}
%
%
%
%
\section{Proposed Dataset} \label{sec:dataset}
\begin{figure*}[t]
\centering
\begin{tabular}{cc}
\includegraphics[width=0.5\textwidth]{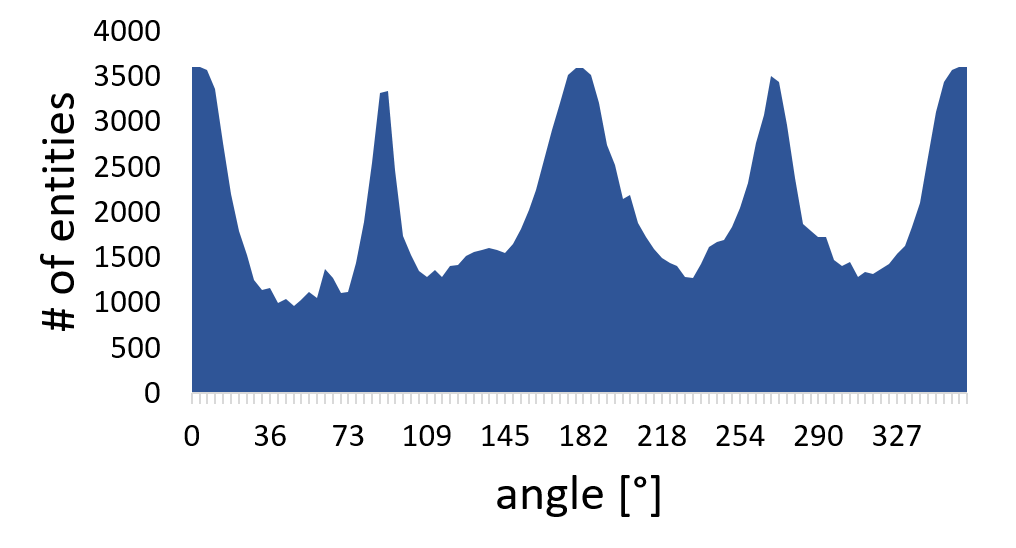}
&
\includegraphics[width=0.5\textwidth]{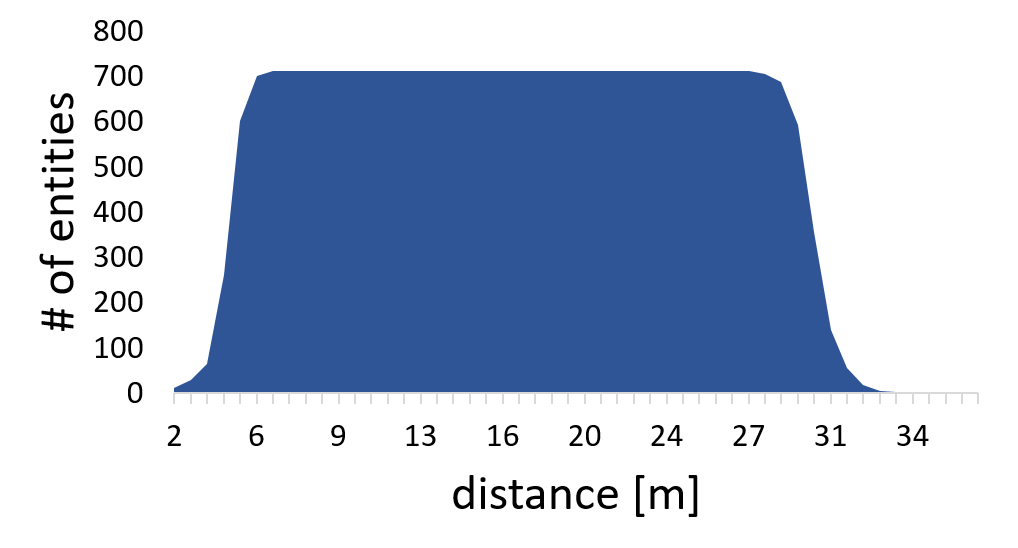}
\\
\hspace{0.7cm}(a)&\hspace{0.45cm}(b)\end{tabular}
\caption{Unnormalized distribution of vehicle orientation (a) and distances (b) present in the collected dataset. Distribution of angles conversely presents two prominent modes around 0$^{\circ}$/360$^{\circ}$ and 180$^{\circ}$ respectively, due to the fact that the major part of vehicles encountered travel in parallel to the player's car, on the same (0/360$^{\circ}$) or the opposite (180$^{\circ}$) direction. Conversely, distance is almost uniformly distributed between $5$ and $30$ meters.}
    \label{fig:angle_and_distance_distribution}
\end{figure*}

In order to collect data, we exploit \emph{Script Hook V} library \cite{synthetic_scripthook}, which allows to use Grand Theft Auto V (GTAV) video game native functions \cite{synthetic_deepgta}.
We develop a framework in which the game camera automatically toggle between frontal and bird-eye view at each game time step: in this way we are able to gather information about the spatial occupancy of the vehicles in the scene from both views (\emph{i.e.} bounding boxes, distances, yaw rotations). We associate vehicles information across the two views by querying the game engine for entity IDs. More formally, for each frame $t$, we compute the set of entities which appear in both views as
\begin{equation}
E(t) = E_{frontal}(t) \cap E_{birdeye}(t)
\end{equation}
where $E_{frontal}(t)$ and $E_{birdeye}(t)$ are the sets of entities that appear at time $t$ in frontal and bird's eye view, respectively. Entities $e(t) \in E(t)$ constitute the candidate set for frame $t$ $C(t)$; other entities are discarded.
Unfortunately, we found that raw data coming from the game engine are not always accurate (Fig. \ref{fig:dataset_samples}). To deal with this problem, we implement a post-processing pipeline in order to discard noisy data from the candidate set $C(t)$. We define a discriminator function
\begin{equation}
f(e(t)) : C \mapsto \lbrace 0, 1 \rbrace
\end{equation}
which is positive when information on dumped data $e(t)$ are reliable and zero otherwise. 
Thus we can define the final filtered dataset as
\begin{equation}
\bigcup_{t=0}^{T} D(t) \qquad \textnormal{where} \quad D(t) = \lbrace c_i(i) \mid f(c_i(t)) > 0\rbrace
\end{equation} 
\noindent being $T$ the total number of frames recorded.
From an implementation standpoint, we employ a rule-based ontology which leverage on entity information (\emph{e.g.} vehicle model, distance etc.) to decide if the bounding box of that entity can be considered reasonable. This implementation has two main values: first it's lightweight and very fast in filtering massive amounts of data. Furthermore, rule parameters can be tuned to eventually generate different dataset distribution (\emph{e.g.} removing all trucks, keeping only cars closer than 10 meters, etc.).\\
Each entry of the dataset is a tuple containing:
\begin{itemize}
\item $frame_{f}$, $frame_{b}$: $1920 \times 1080$ frames from the frontal and bird's eye camera view, respectively;
\item $ID_{e}$, $model_{e}$: identifiers of the entity (e) in the scene and of the vehicle's type;
\item $frontal\_coords_{e}$, $birdeye\_coords_{e}$ : the coordinates of the bounding box that encloses the entity;
\item $distance_{e}$, $yaw_{e}$ : distance and rotation of the entity w.r.t. the player.
\end{itemize}
Fig.~\ref{fig:angle_and_distance_distribution} shows the distributions of entity rotation and distance across the collected data.
\begin{table}[t]
\centering
\begin{tabular}{l | c }
& \quad  Total \\ \hline
Number of runs  & \quad 300 \\
Number of bounding boxes \quad \quad \quad & \quad 1125187 \\
Unique entity IDs & \quad  56454 \\
Unique entity models & \quad  198 \\ \hline
\end{tabular}
\caption{Overview of the statistics on the collected dataset. See text for details.}
\label{tab:dataset_statistics}
\end{table}
%
%
%
\section{Model} \label{sec:model}
\begin{figure*}[t]
\centering
\begin{center}
\includegraphics[width=\textwidth]{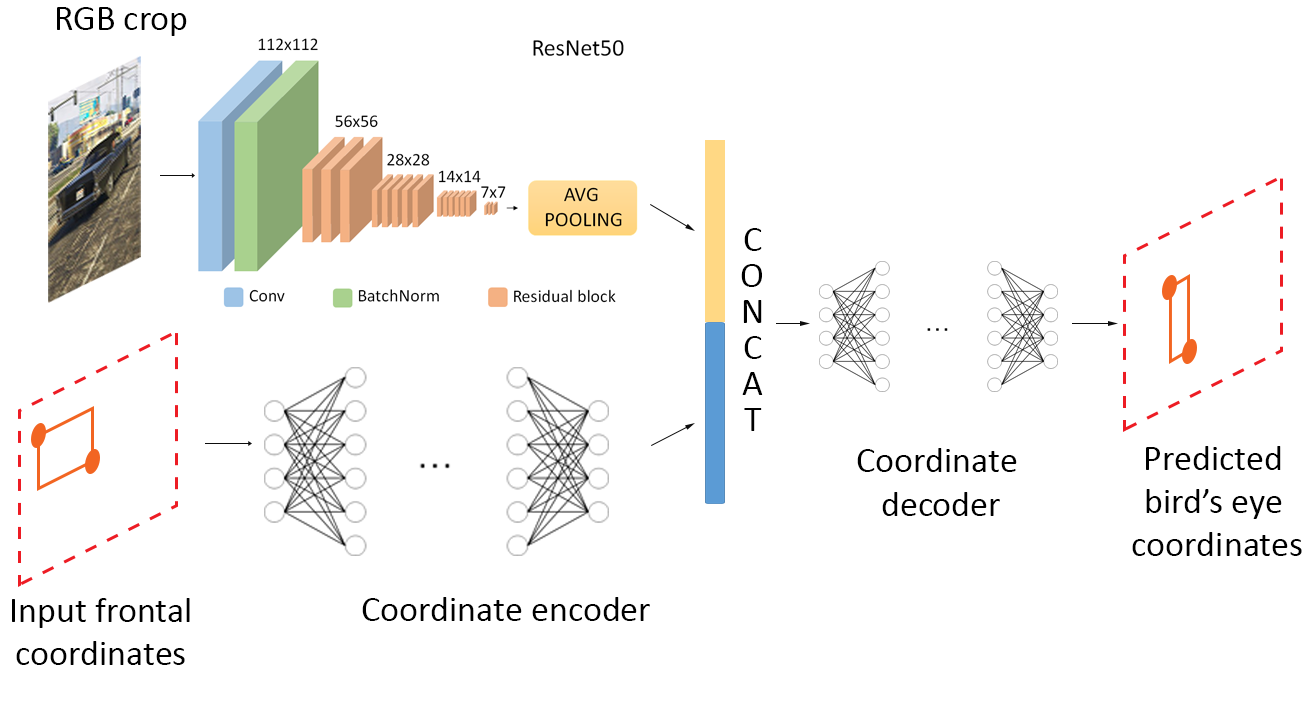}
\end{center}
   \caption{A graphical representation of the proposed \texttt{SDPN} (see Sec. \ref{sec:model}). All layers contain \emph{ReLU} units, except for the top decoder layer which employs \emph{tanh} activation. The number of fully connected units is $(256, 256, 256)$ and $(1024, 1024, 512, 256, 128, 4)$ for the coordinate encoder and decoder respectively.}
\label{fig:model}
\end{figure*}
At a first glance, the problem we address could be mistaken with a bare geometric warping between different views. Indeed, this is not the case since targets are not completely visible from the dashboard camera view and their dimensions in the bird's eye map depend on both the object visual appearance and semantic category (\emph{e.g.} a truck is longer than a car). Additionally, it cannot be cast as a correspondence problem, since no bird's eye view information are available at test time. Conversely, we tackle the problem from a deep learning perspective: dashboard camera information are employed to learn a spatial occupancy map of the scene seen from above.\\ \\
Our proposed architecture composes of two main branches, as depicted in Fig.~\ref{fig:model}. The first branch takes as input image crops of vehicles detected in the dashboard camera view.
We extract deep representations by means of \textit{ResNet50} deep network \cite{he2016deep}, taking advantage of pre-training for image recognition on ImageNet~\cite{deng2009imagenet}. To this end we discard the top fully-connected dense layer which is tailored for the original classification task. This part of the model is able to extract semantic features from input images, even though it is unaware of the location of the bounding box in the scene.\\
Conversely, the second branch consists of a deep \textit{Multi Layer Perceptron} (MLP), composed by 4 fully-connected layers, which is fed with bounding boxes coordinates (4 for each detection), learning to encode the input into a $256$ dimensional feature space. Due to its input domain, this segment of the model is not aware of objects' semantic, and can only learn a spatial transformation between the two planes.\\
Both appearance features and encodings of bounding box coordinates are then merged through concatenation and undergo a further fully-connected decoder which predicts vehicles' locations in the bird's eye view.
Since our model combines information about object's location with semantic hints on the content of the bounding box, we refer to it as \emph{Semantic-aware Dense Projection Network} (\texttt{SDPN} in short).\\ \\
%
\textbf{Training Details:}
ImageNet~\cite{deng2009imagenet} mean pixel value is subtracted from input crops, which are then resized to $224 \times 224$ before being fed to the network. During training, we freeze \emph{ResNet50} parameters. Ground truth coordinates in the bird's eye view are normalized in range $[-1, 1]$. Dropout is applied after each fully-connected layer with drop probability 0.25. The whole model is trained end-to-end using \emph{Mean Squared Error} as objective function and exploiting \textit{Adam}~\cite{DBLP:journals/corr/KingmaB14} optimizer with the following parameters: $lr=0.001, \beta_1=0.9, \beta_2=0.999$.
%
%
%
%
\section{Experimental results}
We now assess our proposal comparing its performance against some baselines. Due to the peculiar nature of the task, the choice of competitor models is not trivial. \\
To validate the choice of a learning perspective against a geometrical one, we introduce a first baseline model that employs a projective transformation to estimate a mapping between corresponding points in the two views. Such correspondences are collected from bottom corners of both source and target boxes in the training set, then used to estimate an homography matrix in a least-squares fashion (\emph{e.g.} minimizing reprojection error). Since correspondences mostly belong to the street, which is a planar region, the choice of the projective transformation seems reasonable. The height of the target box, however, cannot be recovered from the projection, thus it is cast as the average height among training examples. We refer to this model as \emph{homography model}.\\
Additionally, we design second baseline by quantizing spatial locations in both views in a regular grid, and learn point mappings in a probabilistic fashion. For each cell $G^f_i$ in the frontal view grid, a probability distribution is estimated over bird's eye grid cells $G^b_j$, encoding the probability of a pixel belonging to $G^f_i$ to fall in the cell $G^b_j$. During training, top-left and bottom-right bounding box corners in both views are used to update such densities. At prediction stage, given a test point $p_k$ which lies in cell $G^f_i$ we predict destination point by sampling from the corresponding cell distribution. We fix grid resolution to 108x192, meaning a 10x quantization along both axes, and refer to this baseline as \emph{grid model}.\\
It could be questioned if the appearance of the bounding box content in the frontal view is needed at all in estimating the target coordinates, given sufficient training data and an enough powerful model. In order to determine the importance of the visual input in the process of estimating the bird's eye occupancy map, we also train an additional model with approximately the same number of trainable parameters of our proposed model \texttt{SDPN}, but fully connected from input to output coordinates. We refer to this last baseline as \emph{MLP}.\\ \\
\hspace{-0.02cm}
\begin{minipage}{\textwidth}
\begin{minipage}{0.49\textwidth}
\begin{tabular}{r c c c c c}
& IoU $\uparrow$
& CD $\downarrow$
& hE $\downarrow$ 
& wE $\downarrow$
& arE $\downarrow$ \\ \hline
homo & 0.13 & 191.8 & 0.28 & 0.34 & 0.38 \\
grid & 0.18 & 154.3 & 0.74 & 0.70 & 1.30 \\ \hline
MLP & 0.32 & 96.5 & 0.25 & 0.25 & 0.29 \\ 
\texttt{SDPN} & \textbf{0.37} & \textbf{78.0} & \textbf{0.21} & \textbf{0.24} & \textbf{0.29}
\end{tabular}
\end{minipage}
\begin{minipage}{0.5\textwidth}
\includegraphics[width=\textwidth]{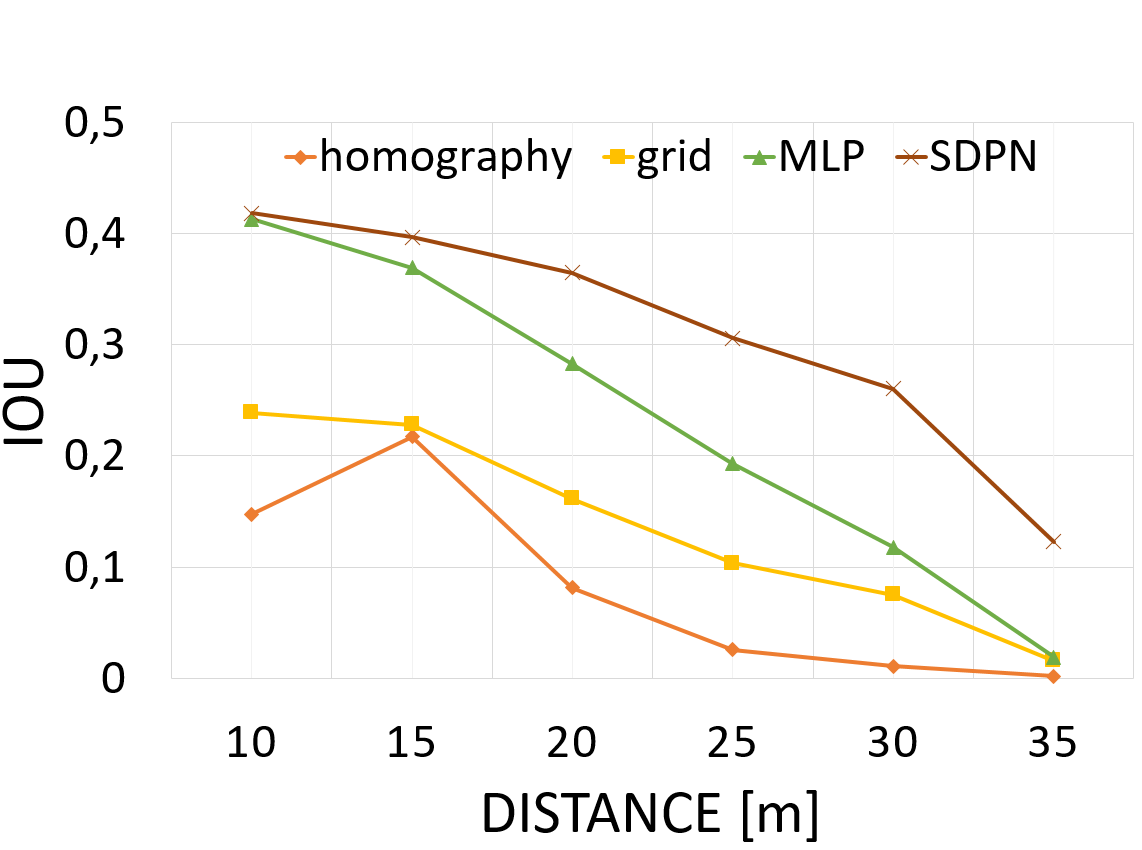}
\end{minipage}
\begin{minipage}{0.5\textwidth}\hspace{2.5cm}(a)\end{minipage}
\begin{minipage}{0.5\textwidth}\hspace{3cm}(b)\end{minipage}
\captionof{figure}{(a) Table summarizing results of proposed \texttt{SDPN} model against the baselines; (b) Degradation of IoU performance as the distance to the detected vehicle increases.}
\label{fig:comparison}
\end{minipage}\\ \\
%
%
For comparison, we rely on three metrics:
\begin{itemize}
\item \textit{Intersection over Union} (IoU): measure of the quality of the predicted bounding box $BB_p$ with respect to the target $BB_t$:
$$IoU(BB_p, BB_t) = \frac{A(BB_p \cap BB_t)}{A(BB_p \cup BB_t)}$$
\noindent where $A(R)$ refers to the area of the rectangle $R$;
\item \textit{Centroid Distance} (CD): distance in pixels between box centers, as an indicator of localization quality\footnote{Please recall that images are 1920x1080 pixel size.};
\item \textit{Height, Width Error} (hE,wE): average error on bounding box height and width respectively, expressed in percentage w.r.t. the ground truth $BB_t$ size;
\item \textit{Aspect ratio mean Error} (arE): absolute difference in aspect ratio between $BB_p$ and $BB_t$:
\begin{equation}
arE= \left \vert \frac{BB_p.w}{BB_p.h} - \frac{BB_t.w} {BB_t.h}  \right \vert
\end{equation}
\end{itemize}
\begin{figure}[tbh]
\centering
\begin{tabular}{ccccc}
\hspace{-0.2cm} \emph{homography} \hspace{1.15cm} & \emph{grid} \hspace{1.5cm} & \emph{MLP} \hspace{1.4cm} & \texttt{SDPN} \hspace{0.8cm} & \emph{ground truth}
\end{tabular}
\begin{center}
\includegraphics[width=\textwidth]{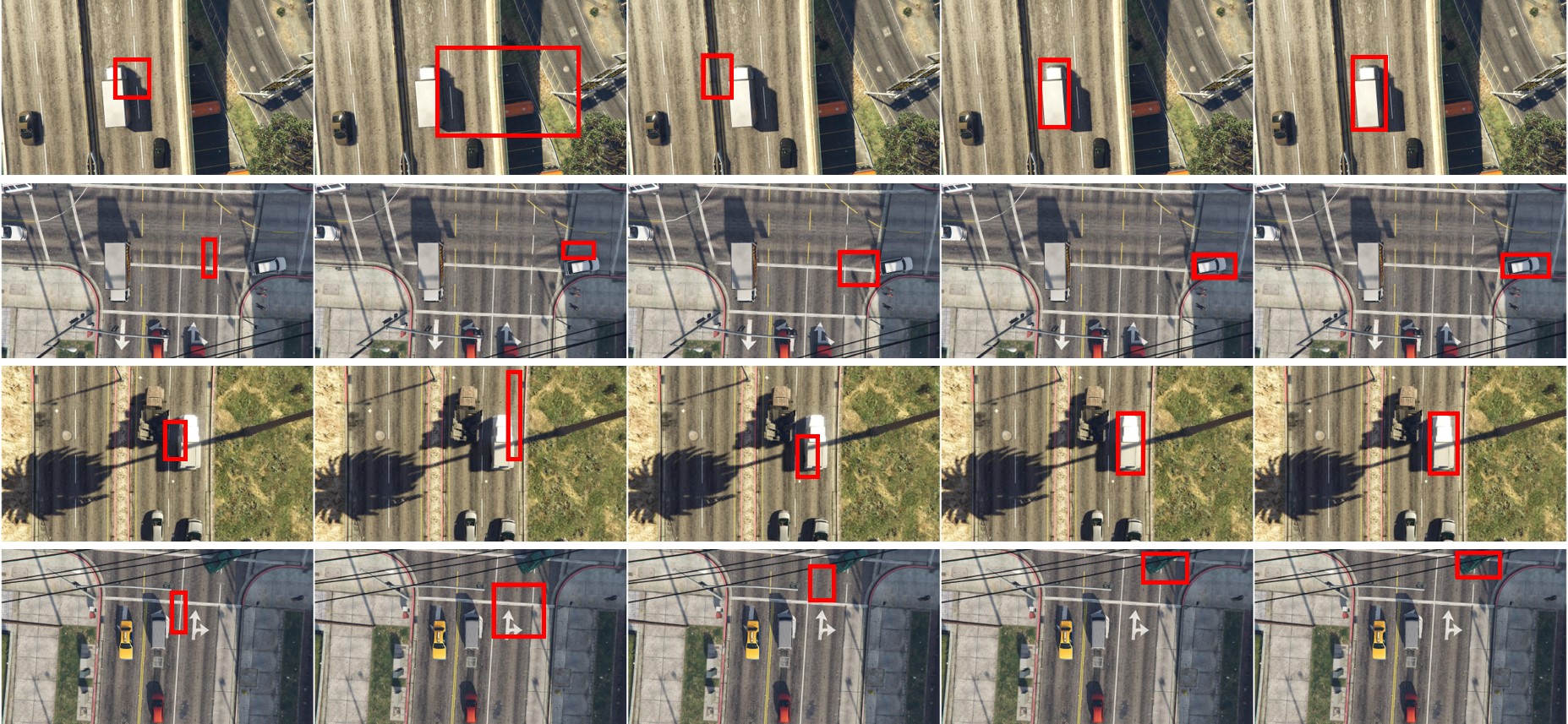}
\end{center}
   \caption{Qualitative comparison between different models. Baselines often predict reasonable locations for the bounding boxes. \texttt{SDPN} is also able to learn the orientation and type of the vehicle (\emph{e.g.} a truck is bigger than a car etc.).}
\label{fig:success_cases}
\end{figure}
\begin{figure}[tbh]
\begin{center}
\includegraphics[width=.9\textwidth]{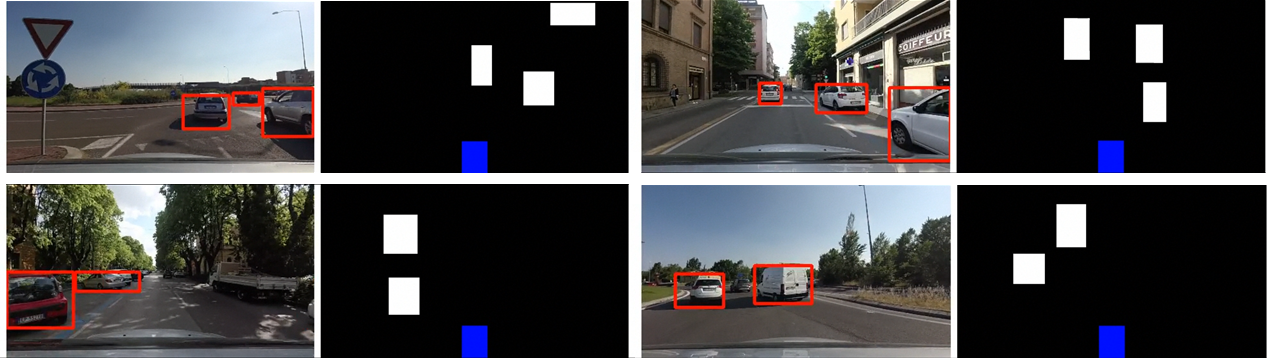}
\end{center}
   \caption{Qualitative results on real-world examples. Predictions look reasonable even if the whole training was conducted on synthetic data.}
\label{fig:realworld}
\end{figure}
The evaluation of baselines and proposed model is reported in Fig.~\ref{fig:comparison}~(a). Results suggest that both \emph{homography} and \emph{grid} are too naive to capture the complexity of the task and fail in properly warping vehicles into the bird's eye view. In particular, \emph{grid} baseline performs poorly as it only models a point-wise transformation between bounding box corners, disregarding information about the overall input bounding box size. On the contrary, MLP processes the bounding box in its whole and provides a reasonable estimation. However, it still misses the chance to properly recover the length of the bounding box in the bird's eye view, being unaware of entity's visual appearance. Instead, \texttt{SDPN} is able to capture the object's semantic, which is a primary cue for correctly inferring vehicle's location and shape in the target view.\\
A second experiment investigates how vehicle's distance affects the warping accuracy. Fig.~\ref{fig:comparison}~(b) highlights that all the models' performance degrades as the distance of target vehicles increases. Indeed, closer examples exhibit lower variance (\emph{e.g.} are mostly related to the car ahead and the ones approaching from the opposite direction) and thus are easier to model. However, it can be noticed that moving forward along distance axis the gap between the \texttt{SDPN} and MLP gets wider. This suggests that the additional visual input adds robustness in these challenging situations. We refer the reader to Fig.~\ref{fig:success_cases} for a qualitative comparison.\\ \\
%
%
\textbf{A real-world case study}
In order to judge the capability of our model to generalize on real-world data, we test it using authentic driving videos taken from a roof-mounted camera~\cite{alletto2016dr}. We rely on state-of-the-art detector~\cite{detection_liu2016ssd} to get the bounding boxes of vehicles in the frontal view. As the ground truth is not available for these sequences, performance is difficult to quantify precisely. Nonetheless, we show qualitative results in Fig. \ref{fig:realworld}: it can be appreciated how the network is able to correctly localize other vehicles' positions, despite having been trained exclusively on synthetic data.\\
\texttt{SDPN} can perform inference at approximately $100Hz$ on a NVIDIA TitanX GPU, which demonstrates the suitability of our model for being integrated in an actual assisted or autonomous driving pipeline.
%
%
%
%
\section{Conclusions} \label{sec:conclusions}
In this paper we presented two main contributions. A new high-quality synthetic dataset, featuring a huge amount of dashboard camera and bird's eye frames, in which the spatial occupancy of a variety of vehicles (i.e. bounding boxes, distance, yaw) is annotated. Furthermore, we presented a deep learning based model to tackle the problem of mapping detections onto a different view of the scene.
We argue that these maps could be useful in an assisted driving context, in order to facilitate driver's decisions by making available in one place a concise representation of the road state. Furthermore, in an autonomous driving scenario, inferred vehicle positions could be integrated with other sensory data such as radar or lidar by means of \emph{e.g.} a Kalman filter to reduce overall uncertainty.

\bibliographystyle{splncs}
\bibliography{biblio}

\end{document}